\documentclass{article}

\usepackage{graphicx}
\usepackage{verbatim}
\usepackage{latexsym}
\usepackage{algorithmic}
\usepackage{amssymb}

\begin{document}

\title{Iterative MILP Methods for Vehicle Control Problems}

\author{Matthew Earl\thanks{Corresponding author. email:
{\tt\small mge1@cornell.edu}}
~and Raffaello D'Andrea}

\date{}
\maketitle

\begin{abstract}
Mixed integer linear programming (MILP) is a powerful tool for
planning and control problems because of its modeling capability and
the availability of good solvers.  However, for large models, MILP
methods suffer computationally.  In this paper, we present iterative
MILP algorithms that address this issue.  We consider trajectory
generation problems with obstacle avoidance requirements and minimum
time trajectory generation problems.  The algorithms use fewer
binary variables than standard MILP methods and require less
computational effort.  
\end{abstract}

\section{Introduction}
Mixed integer linear programming (MILP) methods have attracted
attention because of their modeling capability and because powerful
solvers are available commercially. The utilization of MILP for
modeling and control problems is described in~\cite{Bemporad00} and
for hybrid systems and practical applications in~\cite{Morari03}.
MILP methods are used in~\cite{Ous04a} for cooperative reconnaissance,
in~\cite{Richards02c} for spacecraft path planning, and
in~\cite{Bellingham02,Earl02b,Earl02} for cooperative control
problems. 

Powerful software packages such as CPLEX~\cite{ilog00} solve MILPs
efficiently for problems in which the number of binary variables is of
reasonable size.  However, a major disadvantage of MILP is its
computational complexity.  Because MILP is NP-hard in the number of
binary variables used in the problem formulation~\cite{garey},
computational requirements grow significantly as the number of binary
variables needed to model the problem increases.  Motivated to generate
efficient MILP problem formulations, we have developed several
iterative techniques that require fewer binary variables than standard
MILP methods. 

The MILP obstacle avoidance methods from~\cite{Richards02c} and those
from~\cite{Earl02b,Earl02}, developed independently, specify a
uniformly distributed set of discrete times at which obstacle
avoidance is enforced.  We call this approach uniform gridding.  In
this approach, there is no avoidance guarantee between time steps.
In addition, many of the avoidance times are unnecessary, resulting in
large MILPs that require a significant computational effort to solve.
Here, we present an iterative MILP obstacle avoidance algorithm that
can be used alone or in combination with the uniform gridding
approach.  The algorithm guarantees obstacle avoidance over the entire
trajectory and distributes avoidance times efficiently, resulting in
smaller MILPs that can be solved faster. We also present an iterative
MILP obstacle growing algorithm that allows the use of a coarse set of
uniformly distributed obstacle avoidance times. In this approach,
collision free trajectories are found by artificially increasing the
size of the obstacles that collide with the trajectory generated by
the MILP, iterating until the resulting trajectory is collision free.

Next, we consider the minimum time trajectory generation problem using
MILP.  The MILP approach to this problem presented
in~\cite{Richards02b,Rothwangl01} generates an approximate solution.
Time is discretized uniformly, and an auxiliary binary variable and a
set of inequality constraints are added for each discrete time.  This
approach gives an estimate to the time optimal solution that depends
on the sampling time chosen.  For more accuracy, the sampling time is
reduced, which results in a larger number of binary variables in the
MILP formulation and thus increases the computation time, possibly
exponentially.  Here, we present an iterative MILP algorithm that
solves for the time optimal solution to the problem.  The algorithm
uses binary search. At each iteration, the feasibility of a MILP with
only one discrete time (for the minimum time part of the problem)
needs to be determined. 

The paper is organized as follows: In Section~\ref{vehicle_dynamics},
we describe the dynamics of the vehicles we use to motivate our
methods.  In Section~\ref{sec:avoidance}, we describe two iterative
MILP algorithms for obstacle avoidance, and we perform an average case
computational complexity study comparing the performance of the
iterative time step selection algorithm with the uniform gridding
approach.  Finally, in Section~\ref{mintimeprob}, we describe an
iterative MILP algorithm for minimum time control problems. 
All files for generating the plots found in
this paper are available online~\cite{EarlWebPage}.

\section{Vehicle dynamics}
\label{vehicle_dynamics}
We motivate our methods using the wheeled robots of Cornell's RoboCup
Team~\cite{d'andrea01,stone01}.  In this section, we show how to
simplify their nonlinear governing equations using a procedure
from~\cite{Nagy04}. The result is a linear set of governing equations
coupled by a nonlinear constraint on the control input. This procedure
allows real-time calculation of many near-optimal trajectories and is
a major factor for Cornell's success in the RoboCup
competition.  We then show how to
represent the simplified system in a MILP problem formulation.  The
result is a set of linear discrete time governing equations subject to
a set of linear inequality constraints. 

Each vehicle is equipped with a three-motor omni-directional drive,
which allows it to  move along any direction irrespective of its
orientation.  This allows superior maneuverability compared to
traditional nonholonomic (car-like) vehicles.  
The nondimensional
governing equations of each vehicle are given by
\begin{equation}
  \left[ \begin{array}{c}
    \ddot{x}(t) \\
    \ddot{y}(t) \\
    \ddot{\theta}(t) 
  \end{array} \right] +
  \left[ \begin{array}{c}
    \dot{x}(t) \\
    \dot{y}(t) \\
    \frac{2mL^2}{I}\dot{\theta}(t) 
  \end{array} \right] =
  \mathbf{u}(\theta(t),t),
\end{equation}
where $\mathbf{u}(\theta(t),t) = \mathbf{P}(\theta(t)) \mathbf{U}(t)$,
\begin{eqnarray}
  \mathbf{P}(\theta) = 
  \left[ \begin{array}{ccc}
    -\sin(\theta)&-\sin(\frac{\pi}{3}-\theta)&\sin(\frac{\pi}{3}+\theta) \\ 
    \cos(\theta)&-\cos(\frac{\pi}{3}-\theta)&-\cos(\frac{\pi}{3}+\theta) \\ 
    1 & 1 & 1 
  \end{array} \right]\nonumber,
\end{eqnarray}
and $\mathbf{U}(t) = (U_1(t), U_2(t), U_3(t)) \in \mathcal{U}$.
In these equations $(x(t),y(t))$ are the coordinates of the vehicle,
$\theta(t)$ is its orientation, $\mathbf{u}(\theta(t),t)$ is the
$\theta(t)$-dependent control input, $m$ is the mass of the vehicle,
$I$ is its moment of inertia, $L$ is the distance from the drive to
the center of mass, and $U_i(t)$ is the voltage applied to motor $i$.
The set of admissible voltages $\mathcal{U}$ is the unit cube, and the
set of admissible control inputs is given by $P(\theta) \mathcal{U}$.

These governing equations are coupled and nonlinear.  To
simplify them, we replace the set $P(\theta)\mathcal{U}$ with the
maximal $\theta$-independent set found by taking the intersection of
all possible sets of admissible controls.  The result is a
$\theta$-independent control set defined by control input $(u_x(t), u_y(t),
u_\theta(t))$ and the inequality constraints
$u_x(t)^2 + u_y(t)^2 \leq (3-|u_\theta(t)|)^2/4$ 
and
$|u_\theta(t)| \leq 3$.
Using the restricted set as the allowable control set, the 
governing equations decouple and are given by
\begin{equation}
  \left[ \begin{array}{c}
    \ddot{x}(t) \\
    \ddot{y}(t) \\
    \ddot{\theta}(t) 
  \end{array} \right] +
  \left[ \begin{array}{c}
    \dot{x}(t) \\
    \dot{y}(t) \\
    \frac{2mL^2}{I}\dot{\theta}(t) 
  \end{array} \right] =
  \left[ \begin{array}{c}
    u_x(t) \\
    u_y(t) \\
    u_\theta(t) 
  \end{array} \right].
\end{equation}
The constraints on the control input couple the degrees of freedom.  

To decouple the $\theta$ dynamics we further restrict the admissible
control set to a cylinder defined by the following two inequalities:
$|u_\theta(t)| \leq 1$ and
\setlength{\arraycolsep}{0.0em}
\begin{eqnarray}
\label{nl_u_constraint}
u_x(t)^2 + u_y(t)^2 &{}\leq{}& 1.
\end{eqnarray}
\setlength{\arraycolsep}{5pt}%
Now, the equations of motion for the translational dynamics of the
vehicle are given by
\setlength{\arraycolsep}{0.0em}
\begin{eqnarray}
\ddot{x}(t) + \dot{x}(t) &{}={}& u_x(t),\nonumber\\
\ddot{y}(t) + \dot{y}(t) &{}={}& u_y(t),
\label{eqn:gov}
\end{eqnarray}
\setlength{\arraycolsep}{5pt}%
subject to equation~(\ref{nl_u_constraint}).
In state space form, equation~(\ref{eqn:gov}) is
$\dot{\mathbf{x}}(t) = \mathbf{A}_c \mathbf{x}(t) + \mathbf{B}_c
\mathbf{u}(t)$,
where $\mathbf{x} = (x,y,\dot{x},\dot{y})$ is the state and 
$\mathbf{u} = (u_x,u_y)$ is the control input. 

To represent the governing equations in a MILP framework, we
discretize the control input in time. We require the control input be
constant between time steps.  The result is a set of linear
discrete time governing equations, which we derive next.

Let $N_u$ be the number of discretization steps for the control input
$\mathbf{u}(t)$. Let $t_u[k]$ be the time at step $k$. Let $T_u[k]>0$ be
the time between steps $k$ and $k+1$, for $k \in \{0,\ldots,N_u-1 \}$.
The discrete time governing equations are given by
\setlength{\arraycolsep}{0.0em}
\begin{eqnarray}
\mathbf{x}_u[k+1] &{}={}& \mathbf{A}[k] \mathbf{x}_u[k] 
+ \mathbf{B}[k] \mathbf{u}[k],
\label{disdyn}
\end{eqnarray}
\setlength{\arraycolsep}{5pt}%
where $\mathbf{x}_u[k] = \mathbf{x}(t_u[k])$, 
$\mathbf{u}[k] = \mathbf{u}(t_u[k])$, 
$\mathbf{x}_u[k] = (x_u[k],y_u[k],\dot{x}_u[k],\dot{y}_u[k])$, and
$\mathbf{u}[k] = (u_{x}[k],u_{y}[k])$.  The coefficients
$\mathbf{A}[k]$ and $\mathbf{B}[k]$ are functions of $k$ because we
have allowed for nonuniform time discretizations.  They can be
calculated explicitly in the usual way~\cite{Earl04}.  Because there
will be several different time discretizations used in this paper, we
use subscripts to differentiate them.  In this section, we use the
subscript $u$ to denote variables associated with the discretization
in the control input $\mathbf{u}(t)$.  

The discrete time governing equations can be solved explicitly in the
usual way~\cite{Earl04}.
In later sections of this paper, it will be necessary to represent the
position of the vehicle, at times between control discretization
steps, in terms of the control input.  Because the set of governing
equations is linear, given the discrete state $\mathbf{x}_u[k]$ and the
control input $\mathbf{u}[k]$, we can calculate the vehicle's state 
at any time $t$ using the following equations:
\setlength{\arraycolsep}{0.0em}
\begin{eqnarray}
  x(t)&{}={}& x_u[k] + (1 - e^{t_u[k] - t}) \dot{x}_u[k]\nonumber\\
  &&{+}\:(t - t_u[k] - 1 + e^{t_u[k] - t}) u_{x}[k],\nonumber\\
  \dot{x}(t)&{}={}& (e^{t_u[k] - t}) \dot{x}_u[k]
  + (1 - e^{t_u[k] - t})  u_{x}[k],
  \label{inbetween}
\end{eqnarray}
\setlength{\arraycolsep}{5pt}%
where $k$ satisfies $t_u[k] \leq t \leq t_u[k+1]$.  If the time
discretization of the control input is uniform, $T_u[k_u] = T_u$ for
all $k_u$, then $k_u = \lfloor t/T_u \rfloor$. The components of the
vehicle's state, $y(t)$ and $\dot{y}(t)$, can be calculated in a
similar way.

The control input constraint given by equation~(\ref{nl_u_constraint})
cannot be expressed in a MILP framework because it is nonlinear.  To
incorporate this constraint, we approximate it with a set of linear
inequalities that define a polygon.  The polygon
inscribes the region defined by the nonlinear constraint.  We take the
conservative inscribing polygon to guarantee that the set of allowable
controls defined by the region is feasible.  Similar to work
in~\cite{Richards02b}, we define the polygon by the set of $M_u$
linear inequality constraints 
\begin{eqnarray}
  &&u_{x}[k] \sin \frac{2 \pi m}{M_u} + 
  u_{y}[k] \cos \frac{2 \pi m}{M_u} \leq \cos \frac{\pi}{M_u}\nonumber\\ 
  &&\forall m \in \{ 1,\ldots, M_u \},
  \label{linconstraint}
\end{eqnarray}
for each step $k \in \{ 1,\ldots,N_u \}$.

To illustrate the approach, consider the following minimum control
effort trajectory generation problem. Given a
vehicle governed by equations~(\ref{disdyn})
and~(\ref{linconstraint}), find the sequence of control inputs
$\{\mathbf{u}[k]\}_{k=0}^{N_u-1}$ that transfers the vehicle from
starting state $\mathbf{x}(0) = \mathbf{x}_s$ to finishing state
$\mathbf{x}(t_f) = \mathbf{x}_f$ and minimizes the cost function
\setlength{\arraycolsep}{0.0em}
\begin{eqnarray}
J &{}={}& \sum_{k=0}^{N_u-1} \left( |u_{x}[k]| + |u_{y}[k]| \right).
\label{mincontrolcost1}
\end{eqnarray}
\setlength{\arraycolsep}{5pt}%

To convert the absolute values in the cost function to linear form, we 
introduce auxiliary continuous variables $z_x[k]$ and $z_y[k]$
and the inequality constraints
\setlength{\arraycolsep}{0.0em}
\begin{eqnarray}
&&-z_x[k] \leq u_{x}[k] \leq z_x[k]\nonumber\\
&&-z_y[k] \leq u_{y}[k] \leq z_y[k].
\label{slackconstraints}
\end{eqnarray}
\setlength{\arraycolsep}{5pt}%
Minimizing $z_x[k]$ subject to the inequalities 
$u_{x}[k] \leq z_x[k]$ and
$u_{x}[k] \geq -z_x[k]$ is equivalent to minimizing $|u_x[k]|$
(similarly for $|u_y[k]|)~\cite{Bertsimas97}$.
Using the auxiliary variables, the cost
function can be written as a linear function,
\setlength{\arraycolsep}{0.0em}
\begin{eqnarray}
J &{}={}& \sum_{k=0}^{N_u-1} \left( z_x[k] + z_y[k] \right).
\label{mincontrolcost2}
\end{eqnarray}
\setlength{\arraycolsep}{5pt}%

The resulting optimization problem (minimize~(\ref{mincontrolcost2})
subject
to~(\ref{disdyn}),~(\ref{linconstraint}),~(\ref{slackconstraints}),
and the boundary conditions) is in MILP form.  Because binary
variables do not appear in the problem formulation, it is a linear
program and is easily solved to obtain the optimal sequence of control
inputs.  
\section{Obstacle avoidance}
\label{sec:avoidance}
In vehicle control, it is necessary to avoid other vehicles,
stationary and moving obstacles, and restricted regions.  In this
section, we show how to use MILP to solve obstacle avoidance problems,
we present two iterative MILP obstacle avoidance algorithms that are more
computationally efficient than standard methods, and we perform an
average case computational complexity study.

\subsection{MILP formulation}
We start by showing a MILP method to guarantee circular obstacle
avoidance at $N_o$ discrete times. A version of this method for
uniformly distributed obstacle avoidance times is presented
in~\cite{Richards02c}, and a similar method is presented independently
in~\cite{Earl02b,Earl02}.  The method we present here allows
nonuniform distributions of obstacle avoidance times~\cite{Earl04},
which we take advantage of in our iterative algorithm presented in the
next section.  We use subscript $o$ to denote variables associated
with the time discretization for obstacle avoidance.  For step $k$,
taken to be an element of the set $\{1,\ldots,N_o\}$, let $t_o[k]$ be
the time at which obstacle avoidance is enforced.  Let $R_{obst}$
denote the radius of the obstacle, and let $(x_{obst}[k],y_{obst}[k])$
denote the coordinates of its center at time $t_o[k]$.  We approximate
the obstacle with a polygon, denoted $\mathcal{O}[k]$, defined by a
set of $M_o$ inequalities.  The polygon is given by
\begin{equation} 
\begin{array}{l}
\mathcal{O}[k]=\{ \mbox{ } (\bar{x},\bar{y}) :
(\bar{x}-x_{obst}[k]) \sin \frac{2 \pi m}{M_{o}}\\
\qquad+\:(\bar{y}-y_{obst}[k]) \cos \frac{2 \pi m}{M_{o}}
\leq R_{obst},\\
\qquad\forall m \in \{1,\ldots,M_{o}\} \mbox{ } \}.
\end{array}
\label{obst_con1}
\end{equation}

To guarantee obstacle  avoidance at time $t_o[k]$ the coordinates of
the vehicle must be outside the region $\mathcal{O}[k]$. This
avoidance condition can be written as $(x_o[k],y_o[k]) \notin
\mathcal{O}[k]$, where $(x_o[k],y_o[k])$ are the coordinates of the
vehicle at time $t_o[k]$.  Here $x_o[k] = x(t_o[k])$ and
$y_o[k]=y(t_o[k])$ are expressed in terms of the control inputs using
equation~(\ref{inbetween}).

Because at least one constraint defining the region $\mathcal{O}[k]$
must be violated in order to avoid the obstacle, the avoidance
condition is equivalent to the following condition: there exists an
$m$ such that $(x_o[k]-x_{obst}[k]) \sin \frac{2 \pi
m}{M_{o}}+(y_o[k]-y_{obst}[k]) \cos \frac{2 \pi m}{M_{o}} > R_{obst}$. 

To express this avoidance constraint in a MILP problem formulation, it
must be converted to an equivalent set of linear inequality
constraints. We do so by introducing auxiliary binary variable
$b_m[k] \in \{ 0,1 \}$ and the following $M_o$ inequality constraints:
\begin{equation}
\label{avoidconstraints1}
\begin{array}{l}
(x_o[k]-x_{obst}[k]) \sin \frac{2 \pi m}{M_{o}}
+(y_o[k]-y_{obst}[k]) \cos \frac{2 \pi m}{M_{o}}\\
\qquad> R_{obst} - H b_m[k],
\quad\forall m \in \{ 1,\ldots,M_{o} \}, 
\end{array}
\end{equation}
where $H$ is a large positive number taken to be larger than the
maximum dimension of the vehicle's operating environment plus the
radius of the obstacle.  If $b_m[k] = 1$, the right side of the
inequality is a large, negative number that is always less than the
left side.  In this case, the inequality is inactive because it
is trivially satisfied.  If $b_m[k]= 0$, the inequality is said to be
active because it reduces to an inequality from the existence
condition above.
For obstacle avoidance, at least
one of the constraints in equation~(\ref{avoidconstraints1}) must be
active. To enforce this, we introduce the following inequality
constraint into the problem formulation:
\begin{eqnarray}
  \sum_{m=1}^{M_{o}} b_m[k] \leq M_{o}-1.
  \label{avoidconstraints2}
\end{eqnarray}

Therefore, to enforce obstacle avoidance at time $t_o[k]$, the set of
binary variables $\{b_m[k]\}_{m=1}^{M_o}$ and the constraints given by
equations~(\ref{avoidconstraints1}) and~(\ref{avoidconstraints2}) are
added to the MILP problem formulation.

Consider the example problem from Section~\ref{vehicle_dynamics},
adding obstacles that must be avoided. In this problem, we want to
transfer the vehicle from start state $\mathbf{x}_s$ to finish state
$\mathbf{x}_f$ in time $t_f$ using minimal control effort and avoiding
obstacles.  To enforce obstacle avoidance at each time in the set
$\{t_o[k]\}_{k=1}^{N_o}$, we augment the MILP formulation in
Section~\ref{vehicle_dynamics} with the set of binary variables
$\{b_m[k]\}_{m=1}^{M_o}$, constraints~(\ref{avoidconstraints1}), and
constraint~(\ref{avoidconstraints2}) for all $k$ in the set
$\{1,\ldots,N_o\}$.

\begin{figure}
\centering
\includegraphics[width=180pt]{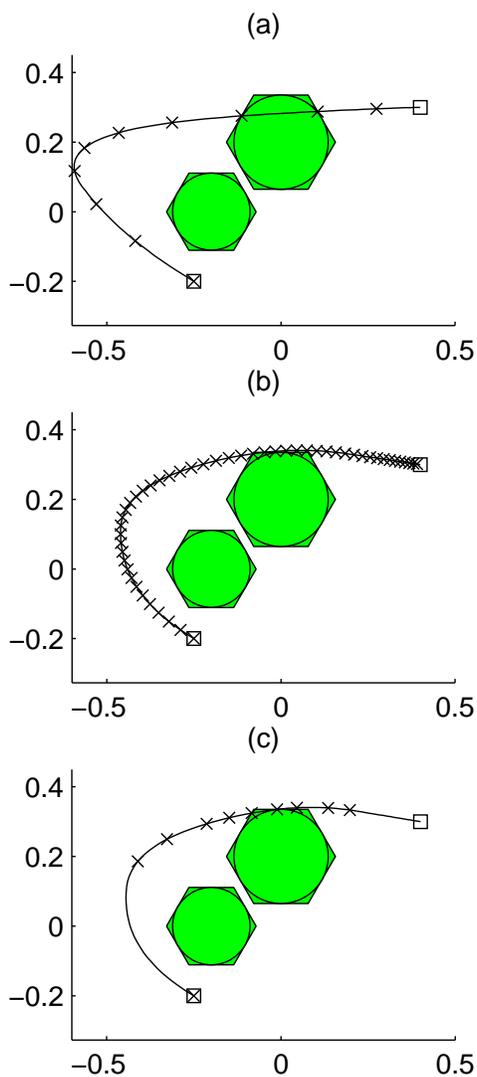}
\caption{Figure~(a) shows the resulting trajectory using uniform
gridding ($N_o=10$), Figure~(b) shows the trajectory using a finer
uniform gridding ($N_o=50$), and Figure~(c) shows the trajectory using
iterative MILP avoidance ($N_o=9$). The circles denote obstacles,
and the polygons denote the buffer regions used in the MILP formulation.
The values of the parameters are $M_o = 6$, $M_u = 4$, $N_u = 4$,
$(x_s,y_s,\dot{x}_s,\dot{y}_s) = (-0.25, -0.2, -0.5, 0.3)$, and
$(x_f,y_f,\dot{x}_f,\dot{y}_f) = (0.4,0.3,0,0)$.}
\label{compareplot}
\end{figure}

Distributing the avoidance times uniformly (uniform gridding) results
in a trajectory that avoids obstacles at each discrete time in the
set. However, the trajectory can collide with obstacles between
avoidance times. This is shown for an example instance in
Figure~\ref{compareplot}(a).

A simple method to reduce this behavior is to take a finer
discretization, which increases the number avoidance times, as shown
in Figure~\ref{compareplot}(b).  However, this is not desirable in
MILP because an increase in the number of avoidance times increases
the number of binary variables in the problem.

\subsection{Iterative MILP time step selection algorithm}
\label{sec:stepSelect}

It is advantageous to use as few avoidance times as possible.
Next, we propose an iterative algorithm to do so.  The method
distributes avoidance times where they are needed most, as shown in
Figure~\ref{compareplot}(c), and guarantees obstacle avoidance if an
obstacle free trajectory exists. The idea is to first solve the MILP
with no obstacle avoidance times (or with a coarse set of avoidance
times) and check the resulting trajectory for collisions. Then, if
there are collisions, augment the MILP formulation with an avoidance
time (and the corresponding binary variables and constraints) for each
collision.  The avoidance time for each collision is taken from the
interval of time that the trajectory is within the obstacle. Next,
solve the augmented MILP and check the resulting trajectory for
collisions, repeating the procedure until a collision free trajectory
is found.
\begin{table}
\caption{Iterative MILP time step selection algorithm}
\label{obstalgo}
\begin{center}
\framebox{\parbox{3.2in}{
\begin{algorithmic}[1]
\STATE Formulate vehicle control problem as a MILP with
the set of obstacle avoidance times $\{t_o[k]\}_{k=1}^{N_o}$.
\STATE Set obstacle buffer zone for each obstacle $j$, 
$R_{buff}^{(j)} := \alpha R_{obst}^{(j)}$ where $\alpha >1$.
\STATE Solve MILP with obstacles of radius $R_{buff}^{(j)}$ for
each obstacle $j$.
\STATE Check resulting trajectory for collisions with obstacles of
radius $R_{obst}^{(j)}$ for each obstacle $j$.
\WHILE {there are collisions}
\STATE For each collision $i$, compute time interval $[t_1^{(i)},t_2^{(i)}]$.
\STATE For each collision $i$, augment the MILP formulation with
       obstacle avoidance constraints at time $t_o^{new} \in
[t_1^{(i)},t_2^{(i)}]$.
\STATE Solve augmented MILP with obstacles of radius $R_{buff}^{(j)}$
for each obstacle $j$.
\STATE Check resulting trajectory for collisions with obstacles of
radius $R_{obst}^{(j)}$ for each obstacle $j$.
\ENDWHILE
\end{algorithmic}
}}
\end{center}
\end{table}

The algorithm is outlined in Table~\ref{obstalgo} and proceeds as
follows: First, formulate the vehicle control problem as a MILP and
choose an initial set of avoidance times $\{t_o[k]\}_{k=1}^{N_o}$.
This set is usually taken to be the empty set or a coarsely
distributed set of times. Next, introduce a buffer zone for
each obstacle $j$ with radius $R_{buff}^{(j)} = \alpha
R_{obst}^{(j)}$, where $\alpha > 1$ is the buffer factor. Radius
$R_{buff}^{(j)}$ is larger than $R_{obst}^{(j)}$ (usually taken
slightly larger) and is used as the radius of obstacle $j$ in the MILP
formulation. This is done to guarantee obstacle avoidance and
termination of the algorithm, which we show later in this section.
Next, solve the MILP using the buffer regions as the obstacles. Then,
check the resulting trajectory for collisions using each obstacle's
true radius, $R_{obst}^{(j)}$ for each obstacle $j$.  To check for
collisions, sample the trajectory and check whether or not each sample
point is inside any of the obstacles.

If there are no collisions, terminate the algorithm.  Otherwise, for
each collision $i$, compute the time interval $[t_1^{(i)},t_2^{(i)}]$
in which the trajectory is within the obstacle.   This interval can be
computed efficiently using a bisection routine and the collision check
routine.  Then, for each collision $i$, augment the MILP problem
formulation with avoidance constraints at time $t_o^{new}$ taken to be
in the interval $[t_1^{(i)},t_2^{(i)}]$. In this paper, we take
$t_o^{new} = (t_1^{(i)}+t_2^{(i)})/2$.  Next, solve the augmented MILP
and check the resulting trajectory for collisions. If there are no
collisions, terminate the algorithm.  Otherwise, repeat the procedure
until there are no collisions.

\begin{figure}
\centering
\includegraphics[width=3.4in]{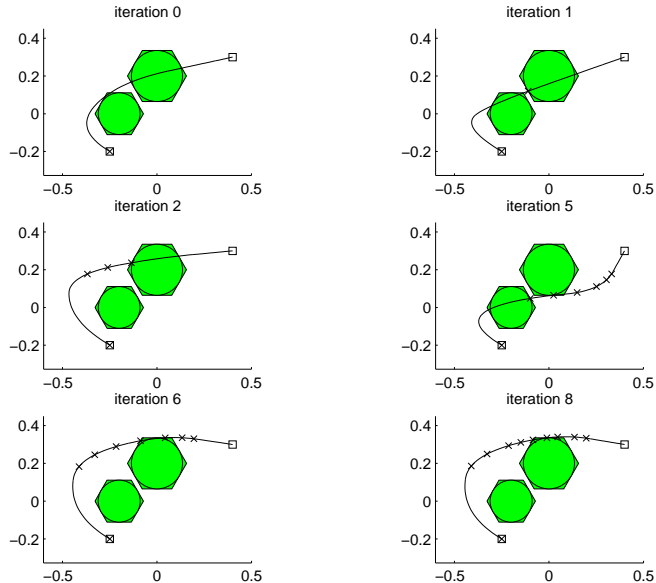}
\caption{Snapshots of the iterative MILP obstacle avoidance algorithm.
The circular regions denote obstacles, and the polygons denote the
buffer regions used in the MILP formulation.  Each cross `$\times$'
denotes a time at which obstacle avoidance is enforced.  The values of
the parameters are $M_o = 6$, $M_u = 4$, $N_u = 4$,
$(x_s,y_s,\dot{x}_s,\dot{y}_s) = (-0.25, -0.2, -0.5, 0.3)$, and
$(x_f,y_f,\dot{x}_f,\dot{y}_f) = (0.4,0.3,0,0)$.
}
\label{actionplot}
\end{figure}

Snapshots of intermediate steps in the iterative algorithm are shown in
Figure~\ref{actionplot}.  The procedure adds obstacle avoidance points
where they are needed most, thus avoiding unnecessary and
computationally costly constraints and binary variables.

\begin{figure}
\centering
\includegraphics[height=80pt]{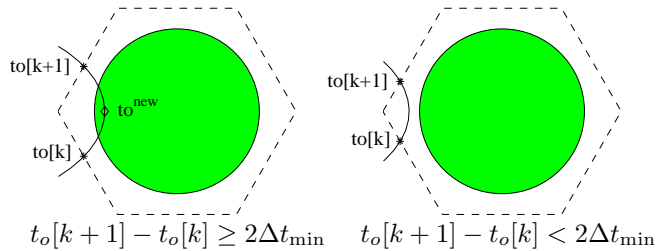}
\put(-225,-10){$t_o[k+1]-t_o[k] \geq 2 \Delta t_{\min}$}
\put(-100,-10){$t_o[k+1]-t_o[k] < 2 \Delta t_{\min}$}
\caption{
These diagrams help show that the iterative MILP time selection
algorithm terminates.  If the difference between consecutive obstacle
avoidance times, denoted $t_o[k+1]-t_o[k]$, is greater than $2 \Delta
t_{\min}$, the trajectory can intersect the obstacle as shown in the
figure on the left. In this case, the algorithm will add a new
avoidance constraint at time $t_o^{new}$. If the difference is less
than $2 \Delta t_{\min}$, the trajectory can not intersect the
obstacle, and a new avoidance constraint can not be added in between.
}
\label{fig:termFig1}
\end{figure}
Now we show that the iterative algorithm in Table~\ref{obstalgo}
terminates.  The minimum distance between the boundary of a buffer
zone and the boundary of the obstacle it surrounds is $d=R_{buff} -
R_{obst} = (\alpha-1)R_{obst}$.  For a problem involving multiple
obstacles, the minimum of these distances is given by $d_{\min} =
(\alpha-1)R_{obst}^{\min}$, where $R_{obst}^{\min}$ is the radius of
the smallest obstacle in the environment.  The minimum time it takes
the vehicle to travel between the boundary of a buffer zone and its
corresponding obstacle is given by $\Delta t_{\min} =
d_{\min}/v_{\max}=(\alpha-1)R_{obst}^{\min}/v_{\max}$, where
$v_{\max}$ is the maximum velocity of the vehicle.  Consider two
consecutive obstacle avoidance times denoted $t_o[k]$ and $t_o[k+1]$.
The vehicle must be located outside all buffer zones at these two
times because we have enforced this as a hard constraint in the MILP.
If the difference $t_o[k+1] - t_o[k]$ is less than $2 \Delta t_{\min}$,
the vehicle's trajectory can not intersect the obstacle because there
is not enough time to enter the buffer zone, collide with the
obstacle, then exit the buffer zone (see Figure~\ref{fig:termFig1}).
In order for the trajectory to intersect the obstacle in the interval
between these two times, the difference $t_o[k+1] - t_o[k]$ must be
greater than $2 \Delta t_{\min}$.  In summary, the algorithm will not
add an obstacle avoidance time in the interval if $t_o[k+1] - t_o[k] <
2 \Delta t_{\min}$, but it can add an obstacle avoidance time if
$t_o[k+1] - t_o[k] \geq 2 \Delta t_{\min}$.  Therefore, in the worst
case, once the algorithm reaches a point where the time interval
between each obstacle avoidance time is less than $2 \Delta t_{\min}$,
the algorithm must terminate.

\begin{figure}
\centering
\includegraphics[width=1.6in]{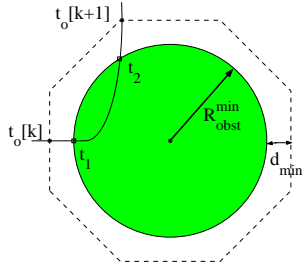}
\caption{
This diagram helps show the minimum time step that can be added by the
iterative MILP time step selection algorithm.  The trajectory
intersects the obstacle in the interval $[t_1,t_2]$.  Assume the
vehicle is moving at $v_{\max}$ between times $t_o[k]$ and $t_1$.  If
the algorithm selects $t_1$ as the new obstacle avoidance time, the
difference $t_1-t_o[k]$ is equal to $\Delta t_{\min}$. This is the
minimum possible time interval between avoidance times because the
vehicle can not move any faster.
}
\label{fig:termFig2}
\end{figure}

Next we bound the number of steps it takes for the algorithm to
terminate. The smallest possible time interval between consecutive
obstacle avoidance times is $\Delta t_{\min}$. This can be seen by
looking at Figure~\ref{fig:termFig2}, where $t_o[k+1]$ and $t_o[k]$
are two consecutive avoidance times and $t_1$ is the time at which the
trajectory enters the obstacle and $t_2$ is the time it exits the
obstacle. Suppose the vehicle is moving at its maximum velocity from
time $t_o[k]$ to $t_1$. The algorithm will detect this intersection,
compute times $t_1$ and $t_2$, and pick a new obstacle avoidance time
$t_o^{new}$ in the interval $[t_1,t_2]$. Suppose the algorithm picks
$t_o^{new} = t_1$, then $t_o^{new} - t_o[k] = \Delta t_{\min}$. The
time interval can not be any less because the vehicle can not pass
through the buffer zone in time less than $\Delta t_{\min}$.  In the
trajectory generation problem, if $t_s$ is the vehicle's starting time
and $t_f$ is its finishing time, the maximum number of time intervals
added by the algorithm is $\lfloor (t_f-t_s)/\Delta t_{\min} \rfloor$.
Therefore, the algorithm will terminate in a maximum of $\lfloor
(t_f-t_s)/\Delta t_{\min} \rfloor$ steps. This is a worst case result.
In practice the algorithm terminates in fewer steps.

\subsection{Iterative MILP obstacle growing algorithm}
Being consistent with our goal to reduce the number of obstacle
avoidance times in our MILP problem formulations, we propose another
iterative MILP algorithm for obstacle avoidance.  This algorithm
iteratively grows the buffer zones surrounding the obstacles until a
collision free trajectory is found.  The idea is to first solve the
MILP with a coarse set of avoidance times and an initial set of buffer
zones surrounding each obstacle. Then, check the resulting trajectory
for collisions. If there are collisions, increase the size of each
buffer zone that surrounds an obstacle with which the trajectory collides.
Next, solve the MILP with these new buffer zones and check the
resulting trajectory for collisions. This process is repeated until
there are no collisions.

\begin{figure}
\centering
\includegraphics[width=3.5in]{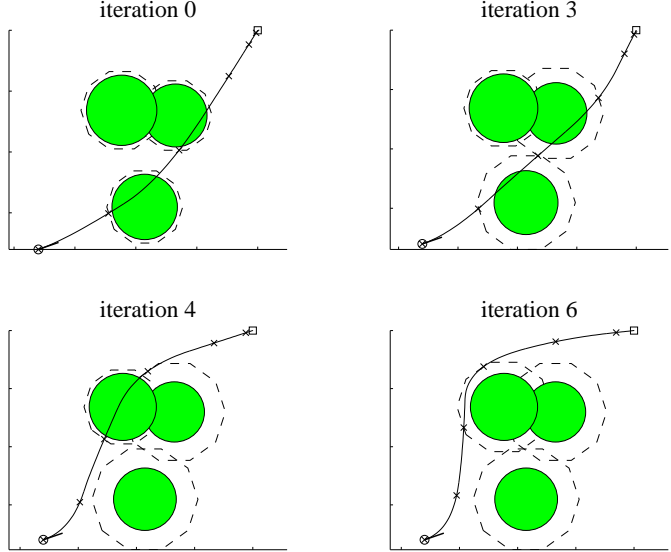}
\caption{Snapshots of the iterative MILP obstacle growing algorithm.
The circular regions denote obstacles, and the polygons denote the
buffer regions used in the MILP formulation. Each cross denotes a
time at which obstacle avoidance is enforced.
}
\label{fig:iterGrow}
\end{figure}

The details of the algorithm are listed in
Table~\ref{tbl:obstGrowAlgo}. Snapshots of intermediate steps of the
algorithm are shown in Figure~\ref{fig:iterGrow}. The crosses denote
the coarse set of times at which obstacle avoidance is enforced in the
MILP. As the figure shows, the size of the buffer regions surrounding
the obstacles with which the trajectory intersects is increased until the
resulting trajectory, generated by the MILP, avoids all obstacles. 

The situation in which this algorithm is most useful is when uniform
gridding is used and the resulting trajectory clips an obstacle, barely
intersecting it. In this case, the algorithm pushes the trajectory
away from the clipped obstacle in a few iterations, resulting in a
collision free trajectory. However, if the initial distribution of
avoidance times is too coarse, the algorithm could have problems. In
this case, the buffer regions could grow to be large and engulf the
initial or final position, which results in an infeasible MILP.

\begin{table}
\caption{Iterative MILP obstacle growing algorithm}
\label{tbl:obstGrowAlgo}
\begin{center}
\framebox{\parbox{3.2in}{
\begin{algorithmic}[1]
\STATE Formulate vehicle control problem as a MILP with
the set of obstacle avoidance times $\{t_o[k]\}_{k=1}^{N_o}$.
\STATE Set obstacle buffer zone for each obstacle $j$, 
$R_{buff}^{(j)} := \alpha R_{obst}^{(j)}$ where $\alpha > 1$.
\STATE Solve MILP with obstacles of radius $R_{buff}^{(j)}$ for
each obstacle $j$.
\STATE Check resulting trajectory for collisions with obstacles of
radius $R_{obst}^{(j)}$ for each obstacle $j$.
\WHILE {there are collisions}
\STATE For each obstacle $j$ that collides with the trajectory,
increase buffer region by setting $R_{buff}^{(j)} := \alpha R_{buff}^{(j)}$.
\STATE Solve MILP with obstacles of radius $R_{buff}^{(j)}$
for each obstacle $j$.
\STATE Check resulting trajectory for collisions with obstacles of
radius $R_{obst}^{(j)}$ for each obstacle $j$.
\ENDWHILE
\end{algorithmic}
}}
\end{center}
\end{table}

\subsection{Average case complexity}
In this section, we explore the average case computational complexity
of the iterative MILP obstacle avoidance algorithm by solving randomly
generated problem instances.  Each instance is generated by randomly
picking parameters from a uniform distribution over the intervals
defined below.  Each MILP is solved using AMPL~\cite{fourer93} and
CPLEX~\cite{ilog00} on a PC with Intel PIII 550MHz processor, 1024KB
cache, 3.8GB RAM, and Red Hat Linux. For all instances solved,
processor speed was the limiting factor, not memory.

For comparison, we solve the same instances using uniform gridding
with sample time $\Delta t_{c}=2R_{obst}^{\min}
\sqrt{\alpha^2-1}/v_{\max}$.
This sample time is the maximum sample time that guarantees obstacle
avoidance, assuming the vehicle travels in a
straight line between sample times. This is a good approximation
since $\Delta t_c$ is small for the instances we solve. See
Appendix~\ref{appendix2} for details.  Each obstacle avoidance time is
given by $t_o[k] = k \Delta t_{c}$, where $k = 1,\ldots,N_o$ and $N_o
= \lceil t_f/\Delta t_{c} \rceil$.

The instances are generated as follows: The start state is taken to be
$\mathbf{x}_s=(x_s, y_s, r_v \cos\theta_v, r_v \sin\theta_v)$, where
$(x_s,y_s)$ is constant, and $r_v$ and $\theta_v$ are random variables
chosen uniformly from the intervals $[r_v^{\min}, r_v^{\max}]$ and
$(0,2\pi]$, respectively. The final state is fixed with zero velocity,
$\mathbf{x}_f = (x_f,y_f,0,0)$. We generate $N_{obst}$ obstacles each
with position $(x_{obst},y_{obst})=(r \cos\theta,r \sin\theta)$ and
radius $R_{obst}$. The parameters $R_{obst}$, $r$, and $\theta$ are
random variables chosen uniformly from the respective intervals
$[R_{\min},R_{\max}]$, $[r_{\min},r_{\max}]$, and $(0,2\pi]$ such that
no obstacle overlaps the circle of radius $R_s$ with position
$(x_s,y_s)$ or the circle of radius $R_f$ with position $(x_f,y_f)$.

For the instances generated in this paper, we set the intervals to be
$r_v \in [0.5,1.0]$, $R_{obst} \in [0.2,0.3]$, and $r \in [0.0,1.0]$.
The constant parameters are taken to be $(x_s,y_s) = (-0.8,-0.8)$,
$(x_f,y_f) = (1.0,1.0)$, $R_s = 0.5$, and $R_f = 0.1$.

\begin{figure}
\centering
\includegraphics[width=3.5in]{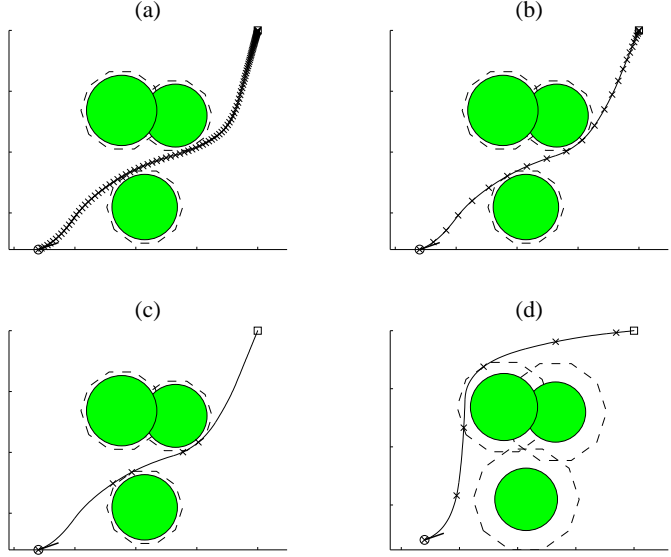}
\caption{
Solutions to an instance of the obstacle avoidance problem using:
(a) uniform gridding with sample time $2 \Delta
t_{\min}$ ($N_o = 115$), (b) uniform gridding with sample time $\Delta
t_c$ ($N_o = 25$), (c) iterative time step selection 
($N_o=4$), and (d) iterative obstacle growing ($N_o = 5$).
The straight line segment denotes the initial velocity $(\dot{x}_s,
\dot{y}_s) = (0.497, 0.172)$, the circular regions denote
obstacles, the dashed regions denote the polygonal buffer zones, and
each cross denotes a time along the trajectory at which obstacle
avoidance is enforced. 
}
\label{fig:3obst}
\end{figure}

The solution to an instance of the obstacle avoidance problem with
three obstacles is shown in Figure~\ref{fig:3obst} for the the uniform
gridding method and for the iterative MILP methods.  Each cross
denotes the time along the trajectory at which obstacle avoidance is
enforced.  The uniform gridding method with sample time $\Delta t_c$
requires $N_o = 25$ obstacle avoidance times, shown in
Figure~\ref{fig:3obst}(b), while the iterative MILP time step
selection algorithm requires only $N_o = 4$ avoidance times, shown in
Figure~\ref{fig:3obst}(c).  Notice the efficiency in which the
iterative algorithm distributes the avoidance times.  For comparison,
we also solve this instance using uniform gridding with sample time $2
\Delta t_{\min}$ (Figure~\ref{fig:3obst}(a)) and using the iterative
obstacle growing algorithm (Figure~\ref{fig:3obst}(d)). For uniform
gridding, choosing sample time $2 \Delta t_{\min}$ guarantees obstacle
avoidance as discussed in Section~\ref{sec:stepSelect}.
However, as shown in the figure, this dense set of obstacle avoidance
times is very conservative.

\begin{figure}
\centering
\includegraphics[width=3.0in]{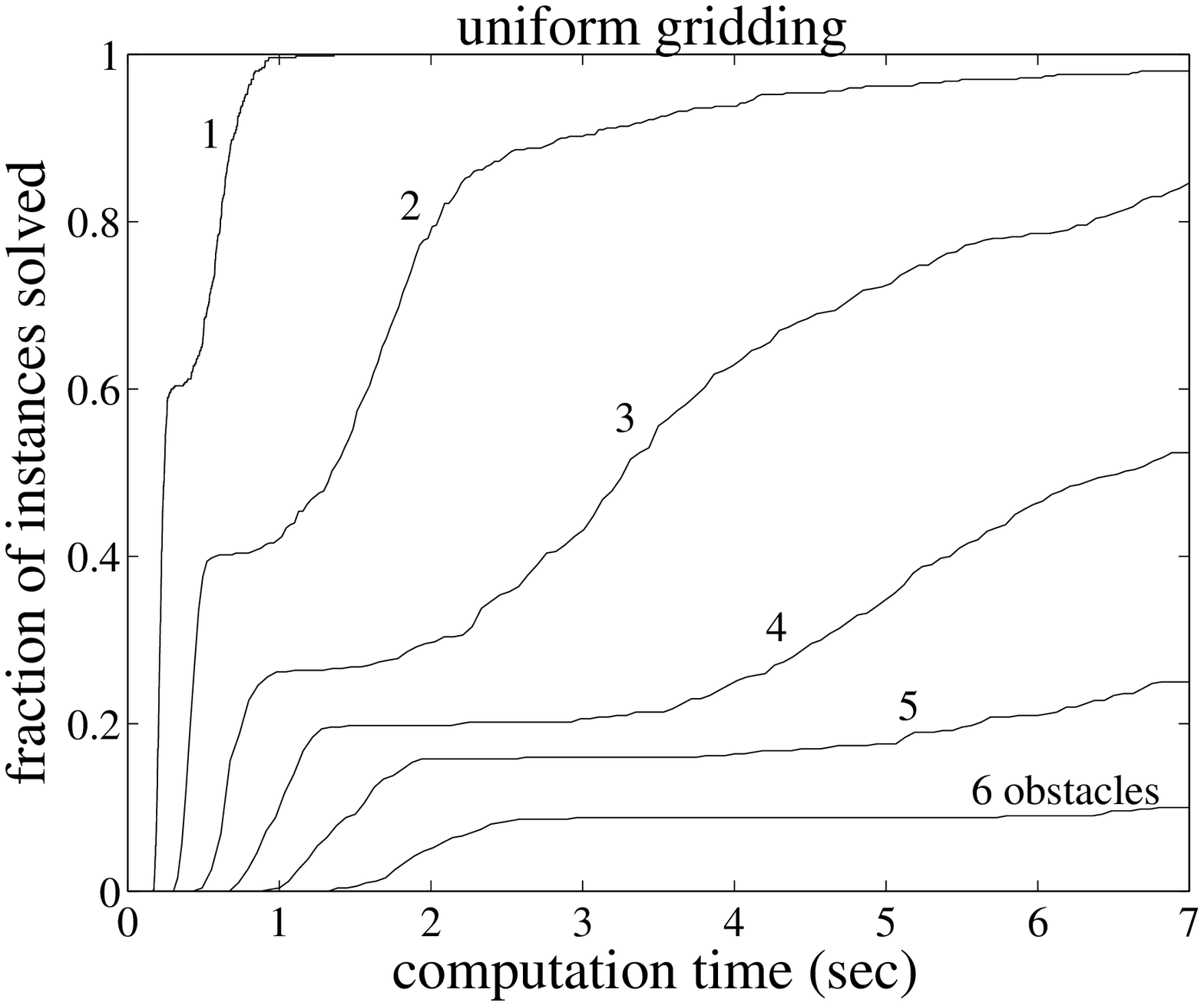}
\includegraphics[width=3.0in]{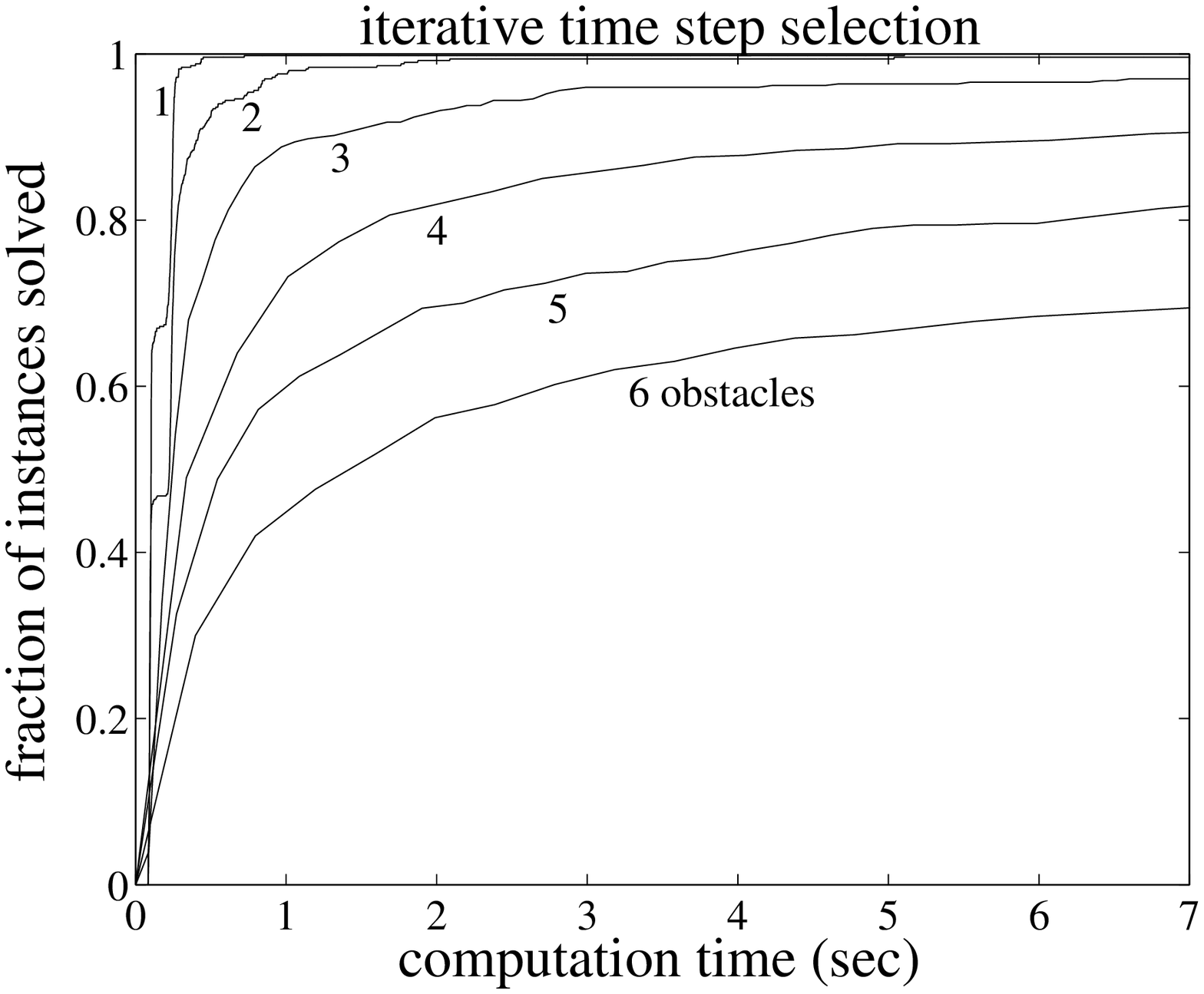}
\caption{Fraction of instances solved versus computation time for
uniform gridding (top) and the iterative MILP obstacle avoidance
(bottom).  We consider $N_{obst} = 2,3,4$. For each curve,
500 random instances were solved.  The values of the parameters
are $N_u = 10$, $M_u = 10$, and $M_o = 10$.}
\label{fig:comptime}
\end{figure}

In Figure~\ref{fig:comptime}, we plot the fraction of instances solved
versus computation time for the two methods. As these figures show,
the iterative MILP method is less computationally intensive than the
uniform gridding method for the instances solved.  For example,
70\% of the instances are solved in 0.4 seconds or less using the
iterative MILP algorithm for the 3 obstacle case. In contrast, no
instances are solved in 0.4 seconds or less using uniform gridding for
the 3 obstacle case.

\begin{figure}
\centering
\includegraphics[width=3.0in]{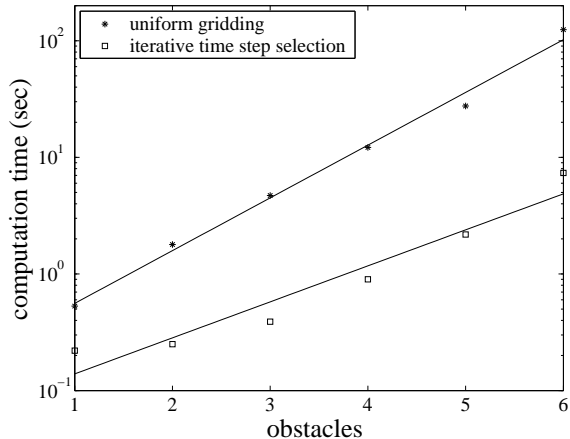}
\caption{Computation time necessary to solve 70\% of the randomly
generated instances versus the number of obstacles.  Each square
denotes a data point for the uniform gridding method.  Each asterisk
denotes a data point for the iterative MILP method.  
The solid lines denote curves fitted to these data.}
\label{fig:expGrow}
\end{figure}

In Figure~\ref{fig:expGrow}, we plot the computation time necessary to
solve 70\% of the randomly generated instances versus the number of
obstacles on the field. Data is plotted for the uniform gridding
method and for the iterative MILP method. The computational
requirements for both methods grow exponentially with the number of
obstacles. However, as the figure shows, the iterative MILP method is
less computationally intensive and the computation time grows at a
slower rate.

\section{Minimum time problems}
\label{mintimeprob}
In this section, we present an iterative MILP algorithm for solving
minimum time problems using a vehicle trajectory generation problem as
motivation.  In~\cite{Richards02b,Rothwangl01}, MILP methods for this
problem are presented.  Time is discretized uniformly and the sampling
interval that contains the optimal time is found using MILP.  To get
better bounds on the optimal time, the sample time of the
discretization must be reduced, which results in a larger number of
binary variables.  In Appendix~\ref{sec:tgprim}, this method is
outlined in the context of the vehicle considered in this paper.

Here we propose an iterative algorithm that converges to the optimal
time using binary search. At each iteration the feasibility of a MILP
is determined using a solver such as CPLEX~\cite{ilog00}. In each MILP, the
number of binary variables and the number of constraints are much fewer 
than those for other techniques because only one discrete time step is
needed.

To motivate the iterative algorithm we consider a minimum time
vehicle control problem. Given a vehicle governed by
equations~(\ref{disdyn}) and~(\ref{linconstraint}), find the sequence of
control inputs $\{ \mathbf{u}[k] \}_{k=0}^{N_u-1}$ that transfers the
vehicle from initial state $\mathbf{x}(0) = \mathbf{x}_s$ to final
state $\mathbf{x}(t_f) = \mathbf{x}_f$ in minimum time.

Suppose we know that the optimal time, denoted $t^*$, is within the time
interval $(t_L,t_R]$. Let time $t_M = (t_L + t_R)/2$.  Consider the
MILP given by equation~(\ref{disdyn}), equation~(\ref{linconstraint}),
constraint $\mathbf{x}(0) = \mathbf{x}_s$, and constraint
$\mathbf{x}(t_f) = \mathbf{x}_f$ with final time taken to be
$t_f=t_M$.  We use equation~(\ref{inbetween}) to express
$\mathbf{x}(t_f)$ in terms of the control inputs. To determine if
there exists a sequence of control inputs that transfers the vehicle
from start state to finish state, we solve the MILP without an
objective function (this is a feasibility problem).

If the MILP is feasible, $t^*$ must be within the interval
$(t_L,t_M]$. Otherwise, the MILP is infeasible and $t^*$ must be
within the interval $(t_M,t_R]$.  By determining the feasibility of
the MILP, we have cut the bound on the optimal time $t^*$ in half.
This suggests an iterative binary search procedure that converges to
$t^*$. 

The iterative algorithm is outlined in Table~\ref{mintimealgo} and
proceeds as follows: First, pick a time interval $(t_{lb},t_{ub}]$
that bounds the optimal time $t^*$.  The lower bound is taken to be
$t_{lb} = d_{\min}/v_{\max}$, where $d_{\min}$ is the straight line
distance from the initial position to the final position and
$v_{\max}$ is the maximum velocity of the vehicle.  The upper bound is
taken to be a feasible time in which the vehicle can reach the
destination.  A simple way to compute a feasible time is to try time
$\alpha t_{lb}$, where $\alpha>1$, increasing $\alpha$ until a
feasible time is found.  Set $t_L := t_{lb}$, $t_R := t_{ub}$, and
$t_M := (t_R + t_L)/2$.

Next, set the final time in the MILP problem formulation to be
$t_f=t_M$, and determine if the resulting MILP is feasible using the
MILP solver. If the MILP is feasible, the optimal time $t^*$ must be
within the interval $(t_L,t_M]$. In this case, set $t_R := t_M$.
Otherwise, the MILP is infeasible and the vehicle can not reach the
destination in time $t_M$.  The optimal time $t^*$ must be within the
interval $(t_M,t_R]$. In this case, set $t_L := t_M$. Then, update
$t_M$ by setting $t_M:=(t_R+t_L)/2$.  If the difference $t_R - t_L$ is
less than some desired tolerance for our calculation of $t^*$, denoted
$\epsilon$, the algorithm terminates.  Otherwise, repeat the process
by setting the final time to $t_f=t_M$ and continue with the steps
outlined previously until the computed value of $t^*$ is within the desired
tolerance $\epsilon$.

After the $k$th iteration, the time interval containing optimal time
$t^*$ has length $(t_{ub} - t_{lb})/2^k$.

\begin{table}
\caption{Iterative minimum time MILP algorithm}
\label{mintimealgo}
\begin{center}
\framebox{\parbox{2.9in}{
\begin{algorithmic}[1]
\STATE Formulate problem as a MILP without objective function.
\STATE Set $t_L := t_{lb}$ and $t_R := t_{ub}$.
\STATE Set $t_M := (t_R + t_L)/2$. 
\WHILE {$(t_R - t_L) > \epsilon$}
\STATE Determine feasibility of MILP with final time $t_f=t_M$.
\STATE {\bf if} feasible  {\bf then} set $t_R := t_M$. 
\STATE {\bf else} set $t_L := t_M$. 
\STATE Set $t_M := (t_R+t_L)/2$. 
\ENDWHILE
\end{algorithmic}
}}
\end{center}
\end{table}

\begin{figure}
\centering
\includegraphics[width=3.3in]{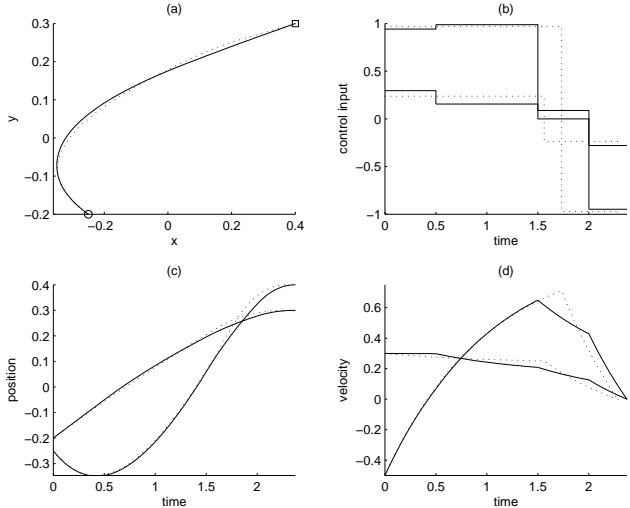}
\caption{Time optimal solution (solid lines) to an instance of the
minimum time vehicle control problem of Section~\ref{mintimeprob}
given by the iterative MILP algorithm.  For comparison we plot the
near optimal solution (dotted lines) for the continuous time version
of the problem obtained using techniques from~\cite{Nagy04}.  The
parameters are: $M_u = 20$, $N_u=10$, $(x_s,y_s,\dot{x}_s,\dot{y}_s) =
(-0.25,-0.2,-0.5,0.3)$, $(x_f,y_f,\dot{x}_f,\dot{y}_f) =
(0.4,0.3,0,0)$.  
}
\label{mintime1}
\end{figure}

The solid lines of Figure~\ref{mintime1} show the solution to an
instance of the minimum time problem.  The iterative procedure was
stopped after thirteen iterations, which took approximately one second
on our Pentium III 550 MHz computer.  To achieve the same accuracy
using the uniform time discretization method, solving one large MILP
with a small sampling time, it took five minutes on the same
computer.  

Our iterative procedure converges to the time optimal solution of the
problem stated in the beginning of this section. This solution is an
approximate solution to the continuous time version of the minimum time
vehicle control problem.  In the continuous time version of the problem,
the vehicle is governed by equations~(\ref{eqn:gov})
and~(\ref{nl_u_constraint}).  We wish to transfer the vehicle from
starting state  $\mathbf{x}_s$ to finishing state $\mathbf{x}_f$ in
minimum time.  In Figure~\ref{mintime1}, we compare our near optimal
solution to the continuous time problem (solid lines) to another
technique (dotted lines) for generating near optimal solutions
from~\cite{Nagy04}, which was used successfully in the RoboCup
competition. 

In addition to being used on its own, our iterative approach can be
combined with the uniform discretization approach. In this case, the
uniform approach is run first with a coarse discretization (large
sampling time $T$). The output is a time interval of size $T$, which
contains the optimal time $t^*$. We use this time interval as the
input to our iterative algorithm.  The $k$th step of the iterative
algorithm outputs a time interval of length $T/2^k$ containing the
optimal time $t^*$, and thus quickly converges to the optimal time. 

\section{Discussion}
We have presented iterative MILP algorithms for obstacle avoidance and
for minimum time control problems.  The iterative MILP time selection
algorithm picks obstacle avoidance times and intelligently distributes
them where they are needed most.  The iterative MILP obstacle growing
algorithm allows a course set of obstacle avoidance times to be used
instead of a dense distribution, which is required to guarantee
obstacle avoidance for standard MILP methods. Both of these algorithms
reduce the number of binary variables needed to formulate and solve
obstacle avoidance trajectory generation problems using MILP.  To
demonstrate the computational benefits of the iterative MILP
time step selection algorithm, we performed an average case
computational complexity analysis.  For comparison, we also performed
the analysis on the standard uniform gridding method.  The iterative
algorithm significantly outperformed the uniform gridding method.  In
addition, we also present an iterative algorithm for solving minimum
time problems using MILP. We found that the algorithm significantly
outperforms standard techniques for minimum time problems using MILP. 

Due to the reduced computational requirements of these
methods, they can be applied more widely in practice.
Computational efficiency is especially important for real
time control in dynamically changing environments where new
control plans need to be generated often and in real time
using a strategy such as model predictive
control~\cite{Mayne00}.  In our
research~\cite{Earl04,Earl04b},  we use these methods to
solve cooperative control problems such as those described
in~\cite{campbell03,d'andrea03a, d'andrea03b}.  However,
there is much room for improvement, including  decreasing
computation time further and developing methods that scale
better with increased numbers of obstacles and vehicles.
In~\cite{Earl04} we discuss ideas to further decrease the
computational requirements of MILP methods. We feel that
intelligent time step selection methods, such as those
presented in this paper, can be very useful in reducing
computational requirements and should be pursued further.
One aspect that needs inspection is the intelligent
selection of the discretization for the control input to the
vehicle.

\appendix

\section{Appendix: Sample time}
\label{appendix2}
Here we derive the minimum sample time, denoted $\Delta t_c$, that
guarantees obstacle avoidance between sample times, assuming the
vehicle moves in a straight line path between sample times. This
is a good approximation, because $\Delta t_c$ is small for the problems
we solve.

Let $d$ be the straight line distance the vehicle can travel  
between any two consecutive avoidance times.
The cord of the smallest buffer region that is tangent to the obstacle
it surrounds is denoted the critical cord. The critical cord length is given
by $d_c = 2((R_{buff}^{\min})^2 - (R_{obst}^{\min})^2)^{1/2}=2
R_{obst}^{\min} \sqrt{\alpha^2 -1}$ because
$R_{buff}^{\min}=\alpha R_{obst}^{\min}$.

If $d<d_c$, the vehicle is guaranteed to avoid the obstacle between
avoidance times. If $d \geq d_c$, the vehicle can collide
with the obstacle between avoidance times.
The critical time interval $\Delta t_c$ is given by
\begin{eqnarray}
\Delta t_c = \frac{d_c}{v_{\max}}=
\frac{ 2R_{obst}^{\min}\sqrt{\alpha^2-1} }
{v_{\max}},
\end{eqnarray}
where $v_{\max}$ is  the maximum velocity of the vehicle.

\section{Appendix: Minimum time MILP formulation}
\label{sec:tgprim}

Here we consider a minimum time trajectory generation problem.  We are
given a vehicle governed by the discrete time system~(\ref{disdyn})
and subject to the constraints~(\ref{linconstraint}).  The objective
is to find the sequence of control inputs $\{ \mathbf{u}[k]
\}_{k=0}^{N_u-1}$ that transfers the system from the initial state
$\mathbf{x}(0) = \mathbf{x}_s$ to the final state $\mathbf{x}(t_f) =
\mathbf{x}_f$ in minimum time. 

Applying the techniques of~\cite{Richards02b,Rothwangl01}, we
introduce a uniform time discretization with constant
sampling time $T$. The solution of the resulting MILP gives
a feasible time that is within $T$ of the optimal time.

Discretize time into $N_T$ times given by $t_T[k] = kT$, where $k$ is
an element of the set $\{ 1,\ldots,N_T\}$. The discretization must be
chosen so that $t_T[N_T] = N_T T$ is larger than the optimal time.

Next, introduce auxiliary binary variable $\delta[k] \in \{ 0,1 \}$
and the inequality constraints,
\begin{eqnarray}
  &&x(t_T[k]) - x_f \leq H(1-\delta[k])\nonumber\\
  &&x(t_T[k]) - x_f \geq -H(1-\delta[k])\nonumber\\
  &&y(t_T[k]) - y_f \leq H(1-\delta[k])\nonumber\\
  &&y(t_T[k]) - y_f \geq -H(1-\delta[k])\nonumber\\
  &&\dot{x}(t_T[k]) -\dot{x}_f \leq H(1-\delta[k])\nonumber\\
  &&\dot{x}(t_T[k]) -\dot{x}_f \geq -H(1-\delta[k])\nonumber\\
  &&\dot{y}(t_T[k]) -\dot{y}_f \leq H(1-\delta[k])\nonumber\\
  &&\dot{y}(t_T[k]) -\dot{y}_f \geq -H(1-\delta[k]),
  \label{eqn:mincon1}
\end{eqnarray}
for each $k$ in the set $\{1, \ldots, N_T\}$. Here, the state
$\mathbf{x}(t_T[k])$ is written in terms of the control inputs using
equation~(\ref{inbetween}), and $H$ is a large positive constant taken
to be greater than the largest dimension of the operating environment.

If $\delta[k]=0$, every constraint in equation~(\ref{eqn:mincon1}) is
trivially satisfied because, for example, $x(t_T[k]) - x_f$ is always
less than $H$.  Otherwise, $\delta[k]=1$ and the constraints in
equation~(\ref{eqn:mincon1}) enforce the condition $\mathbf{x}(t_T[k]
= \mathbf{x}_f)$. To require that the final condition be satisfied at
only one discrete time $t_T[k]$ the following constraint is
introduced,
\begin{equation}
\sum_{i=1}^{N_T} \delta[i] = 1.
\end{equation}

Finally, we introduce the cost function to be minimized,
\begin{equation}
  J = \sum_{i=1}^{N_T} i \delta[i].
\end{equation}
By minimizing this cost the final state $\mathbf{x}_f$ is reached at
the earliest discrete time, $t_T[k]$, possible.  The output after
solving the resulting MILP is a single $k_{sol}$ such that
$\delta[k_{sol}]=1$. The optimal time is therefore within the interval
$( t_T[k_{sol}-1], t_T[k_{sol}] ]$.


\end{document}